\documentclass[10pt,conference]{IEEEtran}
\usepackage[square,sort,comma,numbers]{natbib}

\ifCLASSINFOpdf
   \usepackage[pdftex]{graphicx}
   \graphicspath{{figs/}}
   \DeclareGraphicsExtensions{.pdf,.jpeg,.png}
\else
   \usepackage[dvips]{graphicx}
   \graphicspath{{../figs/}}
   \DeclareGraphicsExtensions{.eps}
\fi

\usepackage[cmex10]{amsmath}
\usepackage{amsthm}

\usepackage{amssymb}

\usepackage{algorithmic}

\usepackage{array}

\ifCLASSOPTIONcompsoc
  \usepackage[caption=false,font=normalsize,labelfont=sf,textfont=sf]{subfig}
\else
  \usepackage[caption=false,font=footnotesize]{subfig}
\fi

\usepackage{url}

\newcommand\Caption[3]{\caption[#2]{\label{#1}\fontsize{8}{10}\textsc{#2}. \footnotesize{#3}}\vspace{-5.0pt}}
\newcommand\etal[1]{~\textit{et al.}~\cite{#1}}
\DeclareMathOperator*{\argmax}{arg\,max}

\newif\iffinal
\finaltrue

\iffinal
\else
\usepackage[switch]{lineno}
\fi

\begin{document}
    
    \title{Open-set Face Recognition with Neural Ensemble, Maximal Entropy Loss and Feature Augmentation}
    
    \iffinal
        \author{
            \IEEEauthorblockN{Rafael Henrique Vareto}
            \IEEEauthorblockA{
                Department of Computer Science\\
                Universidade Federal de Minas Gerais\\
                Belo Horizonte, \textsc{mg}, Brazil\\
                \texttt{\small rafaelvareto@dcc.ufmg.br}
            }
            \and
            \IEEEauthorblockN{Manuel G\"unther}
            \IEEEauthorblockA{
                Department of Informatics\\
                Universit\"at Z\"urich\\
                Z\"urich, \textsc{zh}, Switzerland\\
                \texttt{\small guenther@ifi.uzh.ch}
            }
            \and
            \IEEEauthorblockN{William Robson Schwartz}
            \IEEEauthorblockA{
                Department of Computer Science\\
                Universidade Federal de Minas Gerais\\
                Belo Horizonte, \textsc{mg}, Brazil\\
                \texttt{\small william@dcc.ufmg.br}
            }
        }
    \else
      \author{Sibgrapi paper ID: 57\\ } 
      \linenumbers
    \fi
    
    \maketitle
    \let\thefootnote\relax\footnote{\textsl{Corresponding author's website: \url{https://www.vareto.com.br}}}
    \begin{abstract}
    Open-set face recognition is a scenario in which biometric systems have incomplete knowledge of all existing subjects.
    This arduous requirement must dismiss irrelevant faces and focus on subjects of interest only. 
    For this reason, this work introduces a novel method that associates an ensemble of compact neural networks with data augmentation at the feature level and an entropy-based cost function.
    Deep neural networks pre-trained on large face datasets serve as the preliminary feature extraction module.
    The neural adapter ensemble consists of binary models trained on original feature representations along with negative synthetic mix-up embeddings, which are adequately handled by the designed open-set loss since they do not belong to any known identity.
    We carry out experiments on well-known \textsc{lfw} and \textsc{ijb-c} datasets where results show that the approach is capable of boosting closed and open-set identification accuracy.
\end{abstract}

    \section{Introduction}
\label{sec:introduction}
Not only are face recognition systems sufficiently advanced nowadays to be used in social networks or photo library tagging, but also a leading mechanism to support governments~\cite{delrio2016automated}, law enforcement agencies, and private companies~\cite{du2022elements}.
Despite the significant recent progress, face recognition remains limited when facing poor image quality, a frequent condition in surveillance and \textsc{cctv} environments.
In addition, few researchers have devoted their efforts to solving problems that either require strong generalization or bounded open space risk, a scenario in which input samples are far from any known class and likely to represent unknown distributions.

Open-set face recognition characterizes the scenario where anonymous individuals, unseen during training and enrollment stages, only come into sight during evaluation time~\cite{scheirer2012toward, guenther2017toward}.
As an illustration, one can think of immigration control at airports taking advantage of automated gates and smart face recognition.
The system is expected to dismiss all law-abiding passengers and alert the security personnel whenever criminal offenders turn up.
However, recent newspaper articles have shown that people being misidentified is not a hypothetical exercise but has actually occurred several times across the United States~\cite{romm2017amazons,hill2020wrongfully}.
To make matters worse, false alarms should be avoided by any means since a system identification error may bias the security approach and mistakenly hold up innocent people in custody~\cite{gunther2020watchlist}.

Some late face recognition systems have tackled low-quality images and major improvements have been made with the introduction of specialized loss functions~\cite{liu2017sphereface,wang2018cosface,deng2019arcface,meng2021magface}.
Deep neural networks (\textsc{dnn}) are typically trained on large datasets of public people before being applied to particular face populations~\cite{du2022elements}.
For this reason, the identification task becomes intrinsically domain-adaptive as none of the individuals selected to train the network composes the gallery set, a collection that only contains subjects of interest, also referred to as a watchlist.
Therefore, they often fail to distinguish whether an input face sample is enrolled in the gallery of known individuals since they cannot foresee the unknown.

Several works explore transfer learning techniques or consist of traditional machine learning algorithms fitted on deep feature representations~\cite{guenther2017toward,panareda2017open,bao2018towards}.
Hashing functions have also been used to solve open-set face recognition tasks~\cite{vinzi2010handbook,dos2014extending,vareto2017ijcb}.
G\"unther\etal{gunther2020watchlist} take an existing pretrained deep backbone and replaces its output classification layer with a Neural Adapter Network (\textsc{nan}).
Other approaches rely on clustering techniques that act as a filtering barrier to unknown samples~\cite{henrydoss2020enhancing,vareto2020ijcb}.
Despite all contributions, the aforementioned methods present unbounded open-space risk~\cite{scheirer2012toward} and are not very well suited for rejecting unknown individuals as generally required in the watchlist context.

Many investigators have examined the advantages of ensembles:
Ma\etal{ma2019lightweight} developed an adaptive-boosting classification framework but did not conduct experiments following open-set protocols.
Choi\etal{choi2019ensemble} combined a collection of deep neural networks with Gabor representations whereas Vareto\etal{vareto2020ijcb} employed a clustering technique to filter out dissimilar candidates before training a compact ensemble of binary models.
The former contains a collection of deep neural networks and, in consequence, presents a high computational complexity in both training and evaluation stages.
Similarly, the latter consists of an online training module that ends up making its use in real-time tasks unfeasible.

Dhamija\etal{dhamija2018objectosphere} noticed that unknown features are generally mapped near known classes and, with that in mind, proposed a novel loss function that maximizes the entropy of non-gallery samples.
Data augmentation is widely adopted to prevent overfitting and strengthen the domain generalization capacity of \textsc{dnn}s~\cite{shorten2019survey}.
In furtherance of modifying data in the feature space, Verma\etal{verma2019manifold} came up with an interpolating strategy to generate new feature representations whereas Li\etal{li2021simple} proposed a stochastic feature augmentation procedure to perturb embeddings with Gaussian noise.
None of the previously mentioned works has been evaluated on face benchmarks containing numerous identities and a limited number of samples per class and, as a consequence, they are not an accurate portrayal of realistic biometric tasks.

In this work, we propose a Neural Adapter Ensemble~(\textsc{nae}) of binary learners to handle unbalanced datasets.
Ensembles are generally employed to reduce variance, minimize modeling bias and then decrease overfitting~\cite{tao2019deep}.
During inference, \textsc{nae} aggregates the scores of each inner model and builds a final ranking of candidates.
Moreover, we introduce a margin-based cost function called Maximal Entropy Loss~(\textsc{mel}) that not only produces more rigorous decision boundaries for known classes, but also increases the entropy for negative training samples. 
Since \textsc{mel} relies upon representative negative samples, we develop an Optimized Mix-Up~(\textsc{omu}) feature augmentation method that synthesizes negative embeddings from feature representations of different subjects enrolled in the gallery set.
The data augmentation contributes to the awareness of unknown identities since including artificial embeddings that are appropriately exploited by the cost function can improve the network's generalization performance~\cite{wong2016understanding}.

We conduct experiments considering three pretrained face recognition networks: \textsc{arcface}~\cite{deng2019arcface}, \textsc{vggface2}~\cite{cao2018vggface2} and \textsc{afffe}~\cite{li2018eclipse}.
Results are obtained on two widely-explored benchmarks, namely Labeled Faces in the Wild (\textsc{lfw})~\cite{huang2008labeled} and \textsc{iarpa} Janus Benchmark C (\textsc{ijb-c})\cite{maze2018ijbc}.
Seeing that \textsc{lfw} was initially designed for the face verification task, we adhere to the open-set protocol designed by Günther\etal{guenther2017toward}.
Contrarily, \textsc{ijb-c} specifies an open-set identification guideline named \textsc{test-4} that determines how face recognition algorithms must be evaluated. 
We optimize the hyperparameters on \textsc{lfw} dataset using \textsc{afffe} backbone as the feature representation module.
Then, the same parameters are employed to a subsequent comparison on \textsc{ijb-c} with \textsc{arcface} and \textsc{vggface2} as feature descriptors to check its robustness to different domains.

\begin{figure}
    \centering
    \includegraphics[width=0.92\columnwidth]{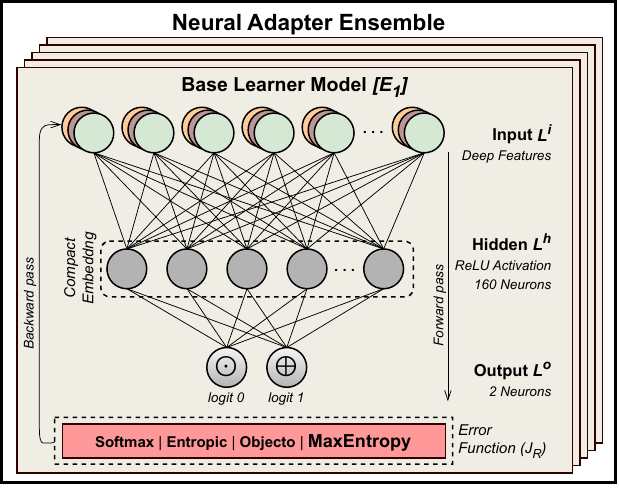}\vspace{-2.0pt}
    \Caption{fig:ensemble}{Neural Ensemble and its Base Learners}{Each standalone learner inputs features extracted with any deep network to learn parameters that minimize an open-set loss function and maps samples from $|G|$ individuals from the gallery set into two classes. For each learner, the ensemble keeps a record of the identities that are randomly associated with class $\odot$ (\textit{output 0}) and class $\oplus$ (\textit{output 1}) to later identify these people and reject unknowns.}
    \vspace{-5.0pt}
\end{figure}

\textbf{The major contributions of our work are:}
\textit{(I)} We present a compact neural ensemble that replaces the computationally-expensive retraining of \textsc{dnn}s for faster ensemble learning.
\textit{(II)} We examine how the entropy-based loss and the feature augmentation mechanism enable the ensemble to better distinguish known from unknown samples.
\textit{(III)} We carry out a parameter selection evaluation to show that the very same setting can be employed in more difficult domains, considering different feature extractors and datasets.
    \section{Proposed Approach}
\label{sec:approach}

A robust open-set face recognition system is expected to determine the identity of those subjects who have been previously enrolled in the gallery and reject the ones of no interest.
However, when deploying biometric applications in the real world, experts must be aware that still and motion probe images very likely present low-quality captures along with illumination variance and occlusion, to name a few~\cite{du2022elements}. 
These corrupted image samples obtained at the testing stage tend to misguide biometric systems and, therefore, end up defiling the identification process.

We design three mechanisms to address the aforestated concerns:
the Neural Adapter Ensemble (\textsc{nae}) encompasses binary neural networks and aims to establish a clear boundary between subjects of interest and unknown faces.
The Optimized Mix-Up~(\textsc{omu}) augmentation synthesizes negative samples at the feature-space level by interpolating representations of different gallery-enrolled individuals.
Maximal-Entropy Loss (\textsc{mel}) comprehends an entropy and margin-based cost function that exploits negative samples derived from the original gallery set or extrinsic datasets.
\textsc{nae} is capable of boosting the predictive performance of each standalone classifier by training multiple base learners and combining their predictions~\cite{zhou2021ensemble}.
\textsc{omu}-made instances can enhance the generalization capability of each binary model that composes the ensemble.
Additionally, \textsc{mel} supports the network in handling unknown samples by maximizing the entropy of negative samples or penalizing the target class of known samples with a specified margin.

\subsection{Feature Extraction}
Given an image sample $x$, the feature extraction module can be defined as $\mathbf{z} = F_\Theta(x)$, a fragment of the pre-trained \textsc{dnn}'s forward pass $\hat{y} = C_\psi \circ F_\Theta(x) = C_\psi \circ F_{\Theta^L} \circ \cdots \circ F_{\Theta^1}(x)$ with classifier $C_\psi$ and $L$ embedding layers in $F_\Theta(x)$. 
This process propagates image $x$ forward up to the point prior to the last fully-connected layer with softmax activation and outputs the equivalent embedding $z$ at that location.
It is important to make sure that the face image $x$ is aligned according to the requirements of the chosen \textsc{dnn}.
Therefore, we rely on pre-determined alignment and feature extraction pipelines~\cite{pereira20228years}.

\subsection{Maximal-Entropy Loss}\label{sub:open-loss}
The Maximal-Entropy Loss (\textsc{mel}) is a cost function that addresses training samples in two different manners:
\textsl{(i)}~\textsc{mel} boosts both intra-class compactness and inter-class separability among known subjects by penalizing the target classes as well as \textsl{(ii)} maximizes the entropy of negative samples by proportionately scattering their output scores among all classes.
\textsc{mel} encloses a soft-margin module~($\mathcal{M}$) with a margin $m \geq 0$ that makes the classification more rigorous.
Then, given a deep feature representation $z$ extracted from a face sample $x$, $s_c(z)$ represents the network activation (logit) for class $c$:
\begin{equation}
    \mathcal{M}_{c}^{m}(z) = \frac{e^{s_c(z)-m}}{e^{s_c(z)-m} + \sum\limits_{c'\neq c}e^{s_{c'}(z)}}
    \label{eq:margin-softmax}
\end{equation}
The formulation of \textsc{mel} ($\mathcal{J}^m$) only adds a handicapping penalty $m$ to known classes, indicated in the first term.
In favor of handling negative samples, \textsc{mel} absorbs the Entropic Open-Set~(\textsc{eos}) loss \cite{dhamija2018objectosphere}, where $y_i$ stores $z_i$'s corresponding target class $c\in C$ and $\bar{z} \in N$ represents a negative sample:
\begin{equation}
    \begin{aligned}
        \mathcal{J}^m = -& \mathbb E_{(z_i, y_i) \in G} \log \mathcal{M}_{y_i}^{m}(z_i)\ \\  
                        -& \mathbb E_{\bar{z} \in N} \frac{1}{|C|} \sum_{c\in C} \log \mathcal{M}_{c}^{m=0}(\bar{z})
    \end{aligned}
    \label{eq:maximal-entropy-loss}
\end{equation}
\textsc{mel} maximizes the uncertainty of negative instances by inducing output activations to lie uniformly distributed over all known classes $c \in C$.
The insight of equalizing logit values for unknown samples lies behind not knowing anything about their corresponding class and, therefore, they must hold a similar likelihood of being assigned to any class~\cite{dhamija2018objectosphere}.
For this reason, $\mathcal{J}^m$ is expected to propagate the entropy learned with negative data $\bar{z}\in N$ to unknown probe samples during inference.

\subsection{Optimized Mix-Up Feature Augmentation}
We introduce an augmentation strategy called Optimized Mix-Up (\textsc{omu}) to build \textsl{artificial} negative samples.
Differently from traditional data augmentation transforms, the designed data synthesis takes place directly at the latent space $z$ and aims to generate tightened decision boundaries around known classes.
\textsc{omu} interpolates two latent embedding representations $z_i$ and $z_j$ into a new representation $\bar{z}$, which is assigned to the negative set $N$. 
Such embedding is generated in consonance with a mingling coefficient $\lambda$ that determines the weight of each original embedding:

\begin{equation}
    \begin{aligned}
    \bar{z} = \lambda \cdot z_i ~+~ (1 - \lambda) \cdot z_j
    \\
    \text{s.t.~}z_j = \argmax_{(z_{i'},g_{i'})\in G}~\cos(z_i, z_{i'}) ~\wedge~ {g_i \neq g_{i'}}
    \end{aligned}
    \label{eq:feature-augmentation-b}
\end{equation}
In summary, given feature vectors $z_i$ and $z_j$, respectively associated with identities $g_i\neq g_j$, a synthetic negative feature $\bar{z}$ is manufactured in between the closest pairs of known individuals.
Unlike existing works~\cite{verma2019manifold,zhou2021learning} where different feature embeddings are randomly selected, \textsc{omu} seeks the closest cosine-similar representations that, at the same time, belong to different subjects registered in gallery $G$.

\subsection{Neural Ensemble Models}
The ensemble is composed of multiple binary classifiers $E_n \in E$.
Each classifier is trained on a different random bisecting split of gallery identities, where the task is to discern these two random groups.
For training our base model $E_n$, we distribute the identities registered in gallery $G$ into two equally-sized disjoint splits.
The random segregation guarantees that half of the individuals are assigned to $P_n^{\odot}$ (partition zero) and the other fraction is allocated in $P_n^{\oplus}$ (partition one) so that both splits altogether, defined as $P_n = \{P_n^{\odot}, P_n^{\oplus}\}$, encompass all the subjects of interest available in the gallery.
Even though all base learners share equivalent architecture and hyperparameters, each one of them is trained with an independent and identically distributed arrangement of known identities as class zero or class one.

Associating subject $g \in G$ with one of the two subsets consists of sampling from a Bernoulli distribution with probability $p=0.5$.
This partitioning $P_n$ operates as the function $B_n : G \mapsto \{\odot,\oplus\}$ that attributes subjects $g$ with new labels ($\odot$ or $\oplus$).
Then, $G\subset\mathbb{Z}_+$ contains the original gallery identities and $B_n$ holds the respective binary co-domain for partition $P_n$.
Theoretically, the probability of any two subjects of interest sharing the very same sequence of binary attributions decreases as the neural ensemble size expands.

The neural ensemble $E$ comprises the main block of the approach since it is the stage in which the feature augmentation scheme and the open-set loss act together to build a set of discriminative base models.
Each base model $E_n \in E$ consists of a multi-layer \textit{perceptron} network with fully-connected layers.
In fact, $E_n$ incorporates an input layer $L^i$, followed by a single hidden layer $L^h$ and an output layer $L^o$.
The input layer takes deep feature representations $z$ extracted with the selected pretrained deep neural network and, consequently, its input size varies according to the \textsc{dnn}'s feature layer dimension. 
As indicated in Figure~\ref{fig:ensemble}, the hidden layer $L^h$ employs a rectified linear unit (\textsc{r}e\textsc{lu}) activation. 
Layer $L^o$ contains two neurons and outputs the corresponding activations $(a^{\odot}, a^{\oplus})$ of the two classes.
Each base learner is trained using \eqref{eq:maximal-entropy-loss}, where we use the categorical loss for two classes.

\begin{table*}[t!]
    \vspace{-10.0pt}
    \Caption{tab:parameters}{LFW Evaluation}{Open-set assessment to select optimal values for parameters $\lambda, h$ and $|E|$.}
    \centering
    \footnotesize
    \begin{tabular}{|c||ccccc|ccccc|ccccc|} \hline 
        \textit{Parameters}             & \multicolumn{5}{c}{$|E|$}                  & \multicolumn{5}{|c|}{$m$}                       & \multicolumn{5}{c|}{$\lambda$}             \\ \hline\hline 
        \textsc{dir}/\textsc{values}    & 0.10 & 0.30 & 0.50    & 0.75     &  1.00   & 0.10    & 0.20    & 0.30    & 0.40    & 0.50    & 0.55    & 0.65    & 0.75    & 0.85    & 0.95     \\ \hline
        $\textsc{tpir@fpir}=1.00$      & 0.82 & 0.92 & \bf0.94 & \bf0.94  & \bf0.94 & \bf0.94 & \bf0.94 & \bf0.94 & \bf0.94 & 0.93    & 0.94    & 0.94    & 0.94    & \bf0.95 & 0.94     \\
        $\textsc{tpir@fpir}=0.10$      & 0.71 & 0.85 & 0.86    & \bf0.88  & \bf0.88 & 0.86    & 0.86    & \bf0.87 & \bf0.87 & \bf0.87 & 0.87    & 0.87    & \bf0.89 & \bf0.89 & 0.87     \\
        $\textsc{tpir@fpir}=0.01$      & 0.57 & 0.72 & 0.73    & \bf0.75  & \bf0.75 & 0.73    & 0.74    & \bf0.77 & 0.76    & 0.74    & \bf0.77 & \bf0.77 & \bf0.77 & 0.74    & 0.75     \\ \hline
    \end{tabular}

\end{table*}

\subsection{Inference with Rank of Candidates}
When given a test sample $x_p$, we first extract the feature embedding $z_p = F_\Theta(x_p)$ and forward these through our ensemble of binary classifiers.
By construction, the activations of the base classifier $E_n$ will be close to zero when an unknown sample is presented, and large for the corresponding partition when facing a known sample~\cite{dhamija2018objectosphere}.
Hence, for each gallery subject $g$, we can simply add the activations of the partition the class $g$ was initially assigned to:
\begin{align}
    \mathrm{sim}(z_p, g) = \sum\limits_{n} a^{B_n(g)}_n(z_p)  
    \label{eq:rank-histogram}
\end{align}
with $B_n(g) \in \{\odot,\oplus\}$. 
The final similarity scores can be used to identify the probe by selecting the gallery subject with the highest score, or rejecting it as unknown when the maximal score is below a certain threshold $\theta$.


    \section{Experiments}
\label{sec:experiments}

We developed the approach using PyTorch framework~\cite{paszke2019pytorch} along with other Python libraries such as Bob \cite{anjos2012bob,pereira20228years} for feature extraction.
The neural ensemble operates on representations obtained with \textsc{senet50-vggface2}~\cite{cao2018vggface2}, \textsc{resnet50-afffe}~\cite{li2018eclipse} and \textsc{resnet100-arcface}~\cite{deng2019arcface} architectures.
Training has been carried out on a dedicated server running Debian Linux on a \textsc{amd epyc 7542} 32-core 128-thread \textsc{cpu}, 512-\textsc{gb ram}, and multiple \textsc{GeForce rtx} 2080Ti \textsc{gpu}s.

\paragraph*{Evaluation Metric}
The open-set Receiver Operating Characteristics (\textsc{o-roc}) is the canonical evaluation metric for open-set biometric systems~\cite{idiap2012beat}.
The \textsc{o-roc} plots the True Positive Identification Rate (\textsc{tpir}) against False Positive Identification Rate (\textsc{fpir}) by varying threshold $\theta$.
The \textsc{tpir} specifies the probability that subjects from the gallery are correctly identified whereas \textsc{fpir} corresponds to the number of unknown subjects mistakenly identified as someone enrolled in the gallery.
An optimal open-set face identification system has \textsc{tpir} of $1$ at an \textsc{fpir} not far from $0$, while the closed-set \textsl{Rank-1} recognition rate can be obtained as \textsc{tpir} @ \textsc{fpir} $=1$.

\paragraph*{Datasets and Evaluation Protocols}
We adopt \textsc{lfw}~\cite{huang2008labeled} as well as \textsc{ijb-c}~\cite{maze2018ijbc} benchmarks.
We incorporate the open-set \textsc{lfw} partitioning~\cite{guenther2017toward} for parameter selection. 
\textsc{ijb-c} provides a widely adopted open-set protocol \textsc{test-4} that consists of two disjoint gallery sets and a probe collection holding identities from both galleries.
We use \textsc{ijb-c}'s \textit{gallery A} for training so that probe samples corresponding to identities available in \textit{gallery B} become unknown.
\textsc{ijb-c} contains mostly high-quality enrollment data but low-quality probe samples, such that the application of simple open-set techniques usually does not transfer from gallery to probes~\cite{linghu2022master}.

\paragraph*{Evaluated Approaches}
We conduct a series of trials in the interest of verifying the enhancement provided by \textsc{nae}, \textsc{mel}, and \textsc{omu}.
We compare the ensemble with \textsc{nan}~\cite{gunther2020watchlist}, a compact network used for multi-class classification. 
We incorporate distinct cost functions in the evaluation stage: angular-based CosFace (\textsc{cfl})~\cite{wang2018cosface}, entropy-maximizing Entropic Open-set (\textsc{eos})~\cite{dhamija2018objectosphere}, and the categorical Cross-Entropy Loss (\textsc{cel}).

\subsection{Parameter Selection}
For an unbiased assessment, we select the parameters obtained on \textsc{lfw} to be subsequently used on \textsc{ijb-c} dataset.
Such practice reveals whether the method is sufficiently robust to generalize across multiple domains and settings.
Table~\ref{tab:parameters} presents results achieved with \textsc{afffe} feature vectors, where we report \textsc{tpir} values for various \textsc{fpir}.
The parametric assessment seeks to find optimal parameters $|E|$, $m$ and $\lambda$, respectively related to the neural ensemble size, loss function penalty, and feature augmentation coefficient.

Empirically, we initially set $|E|=0.1\cdot|G|$, $m=0.1$ and $\lambda=0.55$, where $|G|$ indicates the gallery size with $610$ subjects.
When one of the aforementioned parameters is being modified, the two remaining stand fixed.
After fixing $|E|=0.5\cdot|G|$ as a good compromise between speed and accuracy, we modify $m$ and achieve better model discriminability power for $m=0.3$.
Results show that setting the augmentation coefficient $\lambda=0.85$  (i.e. unknown samples are relatively similar to known samples) increases the ensemble's ability to distinguish subjects of interest from unknowns.
Finally, we assess the optimal number of neurons for hidden layer $L^h$ and, after ranging the number of hidden nodes from $32$ to $256$ in steps of $16$, $L^h$ is empirically set to $160$.

\begin{figure*}[t!]
    \vspace{-3.0pt}
    \centering
    \subfloat[\label{fig:ijbc:na}\textsc{nan} Assessment]{
        \includegraphics[trim={0.3cm 0.5cm 0.3cm 0.85cm},clip,width=.42\textwidth]{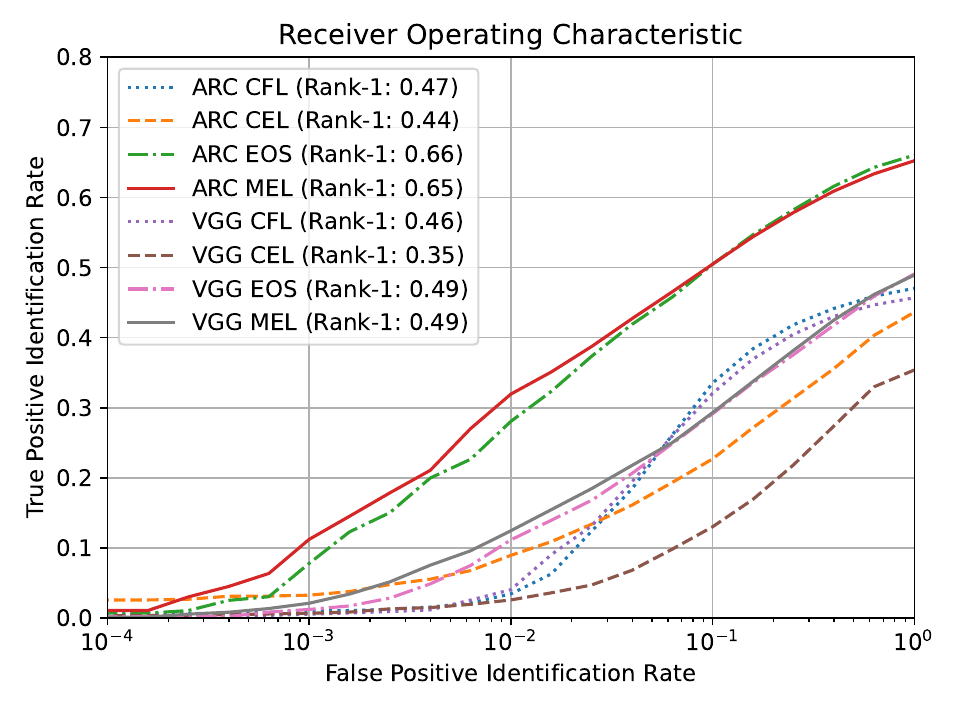}
    }
    \hspace{0.5cm}
    \subfloat[\label{fig:ijbc:ne}\textsc{nae} Assessment]{
        \includegraphics[trim={0.8cm 0.5cm 0.3cm 0.85cm},clip,width=.42\textwidth]{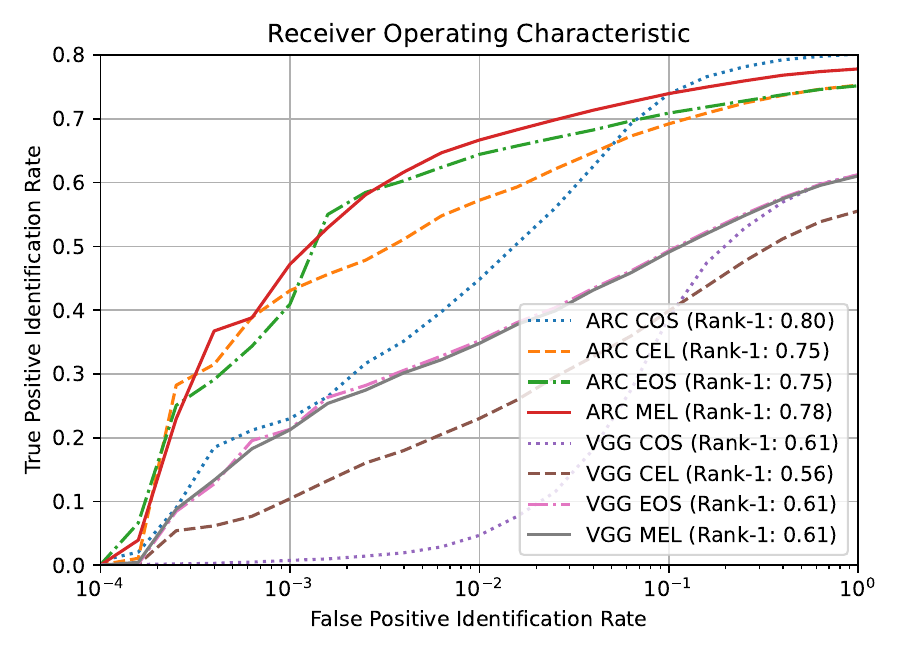}
    }
    \Caption{fig:ijbc}{IJB-C Evaluation}{
        We compare our proposed approach with several state-of-the-art approaches for open-set face recognition, using four different loss functions (\textsc{cel, cfl, eos, mel}) and two deep feature representations \textsc{arcface (arc)} and \textsc{vggface2 (vgg)}.
        \subref{fig:ijbc:na} indicates the improvement obtained with the Neural Adapter Network (\textsc{nan})~\cite{gunther2020watchlist} trained with \textsc{mel} over long-established loss functions, whereas \subref{fig:ijbc:ne} shows the advance brought by Neural Ensemble (\textsc{nae}) over multi-class \textsc{nan} and the well-known cosine-similarity metric (\textsc{cos}) in the open-set identification task.
    }
\end{figure*}

\subsection{Neural Network Evaluation}
The Neural Adapter Network~(\textsc{nan})~\cite{gunther2020watchlist} consists of a multi-layer perceptron with two fully-connected hidden layers.
The first hidden layer encloses 512 neurons with \textsl{ReLU} activation whereas the second one holds 128 neurons.
\textsc{nan} is trained in a multi-class fashion for 200 epochs where the output layer size corresponds to the number of subjects enrolled in the gallery set.
We train the adapter network with four distinct cost functions adhering to the best hyper-parameter specified for margin $m=0.3$ as shown in Table~\ref{tab:parameters}.
The supplementary data is synthesized utilizing the designed \textsc{omu} feature augmentation method with  blending coefficient $\lambda=0.75$.

Figure~\ref{fig:ijbc:na} demonstrates that the addition of synthetic samples holding equivalent underlying data distribution with the gallery set can significantly improve \textsc{nan}'s accuracy.
Not only are the improved results enhanced in the closed-set evaluation (\textsl{Rank-1}) but we can observe the superior performance being propagated to the \textsc{o-roc} metric when the false-positive identification rate decreases.
In other words, \textsc{mel} attains better identification score when \textsc{fpir}$=10^{-3}$ than  \textsc{cel} or \textsc{cfl} lie under \textsc{fpir}$=10^{-2}$.
The chart also demonstrates the advantage of using \textsc{mel} when contrasted with \textsc{eos} as the proposed loss function achieves an analogous closed-set recognition rate but outperforms \textsc{eos} in the open-set evaluation.

\textsc{cfl} is not one of the most recent margin-based cost functions; still, two recent investigations~\cite{kim2022adaface,saadabadi2023quality} demonstrated that it exceeds the results obtained with more sophisticated algorithms on \textsc{ijb-c}, such as ArcFace~\cite{deng2019arcface}, CurricularFace~\cite{huang2020curricularface} and MagFace~\cite{meng2021magface}.
As a result, we believe that CosFace corresponds to a good representation of most angular-margin variants of the traditional Cross-Entropy Loss. 
\textsc{cfl} demands a special parameter setting as it does not operate with probability scores.
During the evaluation of \textsc{cfl}, we had to double the number of epochs as well as the quantity of neurons in the second hidden layer in order to obtain satisfactory results.

\begin{table}[!t]
    \vspace{-3.0pt}
    \centering
    \footnotesize
    \Caption{tab:aug}{Augmentation Analysis}{Evaluation of \textsc{nan} trained with \textsc{cel} or \textsc{mel} associated with different augmentation schemes on \textsc{ijb-c}.}
    \begin{tabular}{|l||cccc|} \hline
        \multicolumn{5}{|c|}{Detection and Identification Rate (\textsc{tpir@})}                            \\ 
        \textsl{Method}       & $\textsc{fpir=}1$ & $\textsc{fpir=}0.1$ & $\textsc{fpir=}0.01$ & $\textsc{fpir=}0.001$ \\ \hline\hline
        \textsc{cel}          & 0.44              & 0.23                & 0.09                 & 0.03  \\ 
        \textsc{mel+lfw}      & 0.58              & 0.23                & 0.10                 & 0.03  \\
        \textsc{mel+sfa}      & \textbf{0.68}     & \textbf{0.53}       & 0.31                 & 0.05  \\
        \textsc{mel+mmu}      & \textbf{0.68}     & 0.52                & 0.28                 & 0.04  \\
        \textsc{mel+omu}      & 0.66              & 0.51                & \textbf{0.33}        & \textbf{0.10}  \\ \hline
    \end{tabular}
    \vspace{-7.0pt}
\end{table}

\paragraph*{Feature Augmentation Analysis}
With \textsc{omu}'s optimal blending parameter $\lambda=0.75$ at hand, we conduct a small set of experiments on \textsc{nan} with \textsc{arcface} in order to check how much improvement can be obtained with the novel augmentation scheme in multi-class tasks.
Table~\ref{tab:aug} compares the proposed augmentation strategy with the Manifold Mix-Up (\textsc{mmu})~\cite{verma2019manifold}, Stochastic Feature Augmentation (\textsc{sfa})~\cite{li2021simple}, and the addition of original \textsc{lfw} samples as the negative set.
Results show that \textsc{cel} cannot keep up with cost functions that explore negative samples.
We observe that either \textsc{sfa} or \textsc{mmu} achieves higher \textsc{tpir} values under higher false-positive identifications, that is, when more unknown samples are mistakenly identified as a gallery-enrolled subject.
\textsc{omu}, however, is capable of achieving better detection and identification rate when \textsc{fpir} drops.

\subsection{Neural Ensemble Evaluation}
Figure~\ref{fig:ijbc} provides a comprehensive comparison between \textsc{nan} and \textsc{nae} as both approaches are trained with analogous feature representations (\textsc{arcface} and \textsc{vggface2}) and cost functions (\textsc{cel}, \textsc{eos} and \textsc{mel}).
Results show the dominant generalization power of ensembles when compared to multi-class models.
We also adopt \textsc{cos}, an abbreviation for the cosine-similarity computation between probe samples and the gallery of templates, as our second baseline.
\textsc{cos} is a common similarity metric for watchlist tasks and is widely employed in the most modern face recognition applications.

As exposed in Figure~\ref{fig:ensemble}, the ensemble consists of multiple binary models containing a single hidden layer with 160 neurons and \textsl{ReLU} activation.
\textsc{nae} exploits synthetic negative samples derived from the gallery set when combined with either \textsc{eos} or \textsc{mel}.
This strategy guarantees a closer data distribution between original and artificially-made training data since 
\textsc{omu} performs a combination of the two closest gallery samples carrying different target classes.
Figure~\ref{fig:ijbc:ne} shows the experimental evaluation of \textsc{cos} and \textsc{nae} on \textsc{ijb-c}.
The top four curves comprise experiments with \textsc{arcface} and the bottom four refer to \textsc{vggface2} feature representations.

The association of \textsc{nae} and \textsc{mel} achieves superior open-set performance when feature representations are extracted with \textsc{arcface} and false-positive identifications range between $10^{-1}$ and $10^{-4}$.
Moreover, \textsc{mel} outperforms \textsc{eos} across many \textsc{fpir} ranges, which shows the importance of learning a more compact feature space through margin $m$.
Experiments with \textsc{vggface2} show that \textsc{eos} and \textsc{mel} cost functions attain competitive results as they outperform methods without negative samples.
Apparently, \textsc{mel} cannot improve the performance over \textsc{eos} as we presume that original \textsc{vggface2} representations do not include sufficient resources to handle \textsc{ijb-c}'s low-quality probe samples and assist \textsc{mel} in the training stage.

Results demonstrate that the proposed approach presents an outstanding performance regardless of the adopted network backbone.
It reveals that supplementary negative data derived from the gallery set itself equips the ensemble with relevant information and boosts the algorithm's overall performance.
In addition, the proposed Maximal-Entropy Loss seems capable of driving each standalone base model toward greater discriminability among known identities as well as escalating the entropy for unknown samples. 
The ensemble acts as an alternate mechanism to prevent the recurrent retraining of very-deep neural networks every time new individuals are enrolled in the gallery set.
As a consequence, it can be attached to the penultimate layer of any pretrained \textsc{dnn}, which eases the maintenance of real-world biometric applications.

    \section{Conclusion}
\label{sec:conclusion}

We introduced three different approaches: a neural ensemble (\textsc{nae}), a cost function (\textsc{mel}), and a feature augmentation algorithm (\textsc{omu}).
Results show that the three methods combined provide better open-set accuracy under the presence of extensive false-positive identifications of unknown samples.
In opposition to most works in the literature, \textsc{nae}, \textsc{mel} and \textsc{omu} did not have their parameters optimized in such a way they would return favorable results on \textsc{ijb-c}.
Instead, the hyper-parameters were selected during the evaluation of a surrogate dataset: \textsc{lfw}.
We believe that the proposed method is likely to achieve higher performance on \textsc{ijb-c} dataset if we take its test set into consideration when tuning the hyper-parameters.

This work also provided an interesting insight: ``transforming a gallery/training set with linear-alike perturbations may provide better generalization capability than external data''.
In fact, one of the experiments showed that synthesizing new samples derived from the gallery set preserves the underlying statistics of the training set and, therefore, ends up contributing more to a model generalization power than extrinsic datasets. 
Consequently, gallery sets with numerous identities but few available samples may not be an obscure limitation anymore when representative synthetic data can be created to assist loss functions in learning better weights.

    \footnotetext{
        \scriptsize
        \textsc{Acknowledgments.} 
        We are thankful to the Brazilian National Council for Scientific and Technological Development -- CNPq (Grants~309953/2019-7 and~203402/2020-0), the Minas Gerais Research Foundation -- FAPEMIG (Grant~PPM-00540-17), the University of Z\"urich (UZH) as well as the Federal University of Minas Gerais (UFMG).
    }

    {
        \footnotesize
        \bibliographystyle{IEEEtran}
        \bibliography{references}
    }

\end{document}